\documentclass[12pt]{article}

\usepackage{sbc-template}
\usepackage{graphicx,url}
\usepackage[utf8]{inputenc}
\usepackage{babel}
\usepackage{lipsum}
\usepackage{footmisc}

\usepackage{changepage}
\usepackage[breakable]{tcolorbox}
    \usepackage{parskip} 
    
    \usepackage{iftex}
    \ifPDFTeX
    	\usepackage[T1]{fontenc}
    	\usepackage{mathpazo}
    \else
    	\usepackage{fontspec}
    \fi

    \usepackage{graphicx}
    
    \usepackage{float}
    \usepackage{xcolor} 
    \usepackage{enumerate} 
    \usepackage{geometry} 
    \usepackage{amsmath} 
    \usepackage{amssymb} 
    \usepackage{textcomp} 
    \AtBeginDocument{%
        \def\PYZsq{\textquotesingle}
    }
    \usepackage{upquote} 
    \usepackage{eurosym} 
    \usepackage[mathletters]{ucs} 
    \usepackage{fancyvrb} 
    \usepackage{grffile} 
    \makeatletter 
    \@ifpackagelater{grffile}{2019/11/01}
    {
    }
    {
      \def\Gread@@xetex#1{%
        \IfFileExists{"\Gin@base".bb}%
        {\Gread@eps{\Gin@base.bb}}%
        {\Gread@@xetex@aux#1}%
      }
    }
    \makeatother
    \usepackage[Export]{adjustbox} 
    \adjustboxset{max size={0.9\linewidth}{0.9\paperheight}}

    \usepackage{hyperref}
    \usepackage{longtable} 
    \usepackage{booktabs}  
    \usepackage[inline]{enumitem} 
    \usepackage[normalem]{ulem} 
    \usepackage{mathrsfs}

    \definecolor{urlcolor}{rgb}{0,.145,.698}
    \definecolor{linkcolor}{rgb}{.71,0.21,0.01}
    \definecolor{citecolor}{rgb}{.12,.54,.11}

    \definecolor{ansi-black}{HTML}{3E424D}
    \definecolor{ansi-black-intense}{HTML}{282C36}
    \definecolor{ansi-red}{HTML}{E75C58}
    \definecolor{ansi-red-intense}{HTML}{B22B31}
    \definecolor{ansi-green}{HTML}{00A250}
    \definecolor{ansi-green-intense}{HTML}{007427}
    \definecolor{ansi-yellow}{HTML}{DDB62B}
    \definecolor{ansi-yellow-intense}{HTML}{B27D12}
    \definecolor{ansi-blue}{HTML}{208FFB}
    \definecolor{ansi-blue-intense}{HTML}{0065CA}
    \definecolor{ansi-magenta}{HTML}{D160C4}
    \definecolor{ansi-magenta-intense}{HTML}{A03196}
    \definecolor{ansi-cyan}{HTML}{60C6C8}
    \definecolor{ansi-cyan-intense}{HTML}{258F8F}
    \definecolor{ansi-white}{HTML}{C5C1B4}
    \definecolor{ansi-white-intense}{HTML}{A1A6B2}
    \definecolor{ansi-default-inverse-fg}{HTML}{FFFFFF}
    \definecolor{ansi-default-inverse-bg}{HTML}{000000}

    \definecolor{outerrorbackground}{HTML}{FFDFDF}

    \DefineVerbatimEnvironment{Highlighting}{Verbatim}{commandchars=\\\{\}}

    \let\Oldtex\TeX
    \let\Oldlatex\LaTeX
    \renewcommand{\TeX}{\textrm{\Oldtex}}
    \renewcommand{\LaTeX}{\textrm{\Oldlatex}}

\makeatletter
\def\PY@reset{\let\PY@it=\relax \let\PY@bf=\relax%
    \let\PY@ul=\relax \let\PY@tc=\relax%
    \let\PY@bc=\relax \let\PY@ff=\relax}
\def\PY@tok#1{\csname PY@tok@#1\endcsname}
\def\PY@toks#1+{\ifx\relax#1\empty\else%
    \PY@tok{#1}\expandafter\PY@toks\fi}
\def\PY@do#1{\PY@bc{\PY@tc{\PY@ul{%
    \PY@it{\PY@bf{\PY@ff{#1}}}}}}}
\def\PY#1#2{\PY@reset\PY@toks#1+\relax+\PY@do{#2}}

\@namedef{PY@tok@w}{\def\PY@tc##1{\textcolor[rgb]{0.73,0.73,0.73}{##1}}}
\@namedef{PY@tok@c}{\let\PY@it=\textit\def\PY@tc##1{\textcolor[rgb]{0.25,0.50,0.50}{##1}}}
\@namedef{PY@tok@cp}{\def\PY@tc##1{\textcolor[rgb]{0.74,0.48,0.00}{##1}}}
\@namedef{PY@tok@k}{\let\PY@bf=\textbf\def\PY@tc##1{\textcolor[rgb]{0.00,0.50,0.00}{##1}}}
\@namedef{PY@tok@kp}{\def\PY@tc##1{\textcolor[rgb]{0.00,0.50,0.00}{##1}}}
\@namedef{PY@tok@kt}{\def\PY@tc##1{\textcolor[rgb]{0.69,0.00,0.25}{##1}}}
\@namedef{PY@tok@o}{\def\PY@tc##1{\textcolor[rgb]{0.40,0.40,0.40}{##1}}}
\@namedef{PY@tok@ow}{\let\PY@bf=\textbf\def\PY@tc##1{\textcolor[rgb]{0.67,0.13,1.00}{##1}}}
\@namedef{PY@tok@nb}{\def\PY@tc##1{\textcolor[rgb]{0.00,0.50,0.00}{##1}}}
\@namedef{PY@tok@nf}{\def\PY@tc##1{\textcolor[rgb]{0.00,0.00,1.00}{##1}}}
\@namedef{PY@tok@nc}{\let\PY@bf=\textbf\def\PY@tc##1{\textcolor[rgb]{0.00,0.00,1.00}{##1}}}
\@namedef{PY@tok@nn}{\let\PY@bf=\textbf\def\PY@tc##1{\textcolor[rgb]{0.00,0.00,1.00}{##1}}}
\@namedef{PY@tok@ne}{\let\PY@bf=\textbf\def\PY@tc##1{\textcolor[rgb]{0.82,0.25,0.23}{##1}}}
\@namedef{PY@tok@nv}{\def\PY@tc##1{\textcolor[rgb]{0.10,0.09,0.49}{##1}}}
\@namedef{PY@tok@no}{\def\PY@tc##1{\textcolor[rgb]{0.53,0.00,0.00}{##1}}}
\@namedef{PY@tok@nl}{\def\PY@tc##1{\textcolor[rgb]{0.63,0.63,0.00}{##1}}}
\@namedef{PY@tok@ni}{\let\PY@bf=\textbf\def\PY@tc##1{\textcolor[rgb]{0.60,0.60,0.60}{##1}}}
\@namedef{PY@tok@na}{\def\PY@tc##1{\textcolor[rgb]{0.49,0.56,0.16}{##1}}}
\@namedef{PY@tok@nt}{\let\PY@bf=\textbf\def\PY@tc##1{\textcolor[rgb]{0.00,0.50,0.00}{##1}}}
\@namedef{PY@tok@nd}{\def\PY@tc##1{\textcolor[rgb]{0.67,0.13,1.00}{##1}}}
\@namedef{PY@tok@s}{\def\PY@tc##1{\textcolor[rgb]{0.73,0.13,0.13}{##1}}}
\@namedef{PY@tok@sd}{\let\PY@it=\textit\def\PY@tc##1{\textcolor[rgb]{0.73,0.13,0.13}{##1}}}
\@namedef{PY@tok@si}{\let\PY@bf=\textbf\def\PY@tc##1{\textcolor[rgb]{0.73,0.40,0.53}{##1}}}
\@namedef{PY@tok@se}{\let\PY@bf=\textbf\def\PY@tc##1{\textcolor[rgb]{0.73,0.40,0.13}{##1}}}
\@namedef{PY@tok@sr}{\def\PY@tc##1{\textcolor[rgb]{0.73,0.40,0.53}{##1}}}
\@namedef{PY@tok@ss}{\def\PY@tc##1{\textcolor[rgb]{0.10,0.09,0.49}{##1}}}
\@namedef{PY@tok@sx}{\def\PY@tc##1{\textcolor[rgb]{0.00,0.50,0.00}{##1}}}
\@namedef{PY@tok@m}{\def\PY@tc##1{\textcolor[rgb]{0.40,0.40,0.40}{##1}}}
\@namedef{PY@tok@gh}{\let\PY@bf=\textbf\def\PY@tc##1{\textcolor[rgb]{0.00,0.00,0.50}{##1}}}
\@namedef{PY@tok@gu}{\let\PY@bf=\textbf\def\PY@tc##1{\textcolor[rgb]{0.50,0.00,0.50}{##1}}}
\@namedef{PY@tok@gd}{\def\PY@tc##1{\textcolor[rgb]{0.63,0.00,0.00}{##1}}}
\@namedef{PY@tok@gi}{\def\PY@tc##1{\textcolor[rgb]{0.00,0.63,0.00}{##1}}}
\@namedef{PY@tok@gr}{\def\PY@tc##1{\textcolor[rgb]{1.00,0.00,0.00}{##1}}}
\@namedef{PY@tok@ge}{\let\PY@it=\textit}
\@namedef{PY@tok@gs}{\let\PY@bf=\textbf}
\@namedef{PY@tok@gp}{\let\PY@bf=\textbf\def\PY@tc##1{\textcolor[rgb]{0.00,0.00,0.50}{##1}}}
\@namedef{PY@tok@go}{\def\PY@tc##1{\textcolor[rgb]{0.53,0.53,0.53}{##1}}}
\@namedef{PY@tok@gt}{\def\PY@tc##1{\textcolor[rgb]{0.00,0.27,0.87}{##1}}}
\@namedef{PY@tok@err}{\def\PY@bc##1{{\setlength{\fboxsep}{\string -\fboxrule}\fcolorbox[rgb]{1.00,0.00,0.00}{1,1,1}{\strut ##1}}}}
\@namedef{PY@tok@kc}{\let\PY@bf=\textbf\def\PY@tc##1{\textcolor[rgb]{0.00,0.50,0.00}{##1}}}
\@namedef{PY@tok@kd}{\let\PY@bf=\textbf\def\PY@tc##1{\textcolor[rgb]{0.00,0.50,0.00}{##1}}}
\@namedef{PY@tok@kn}{\let\PY@bf=\textbf\def\PY@tc##1{\textcolor[rgb]{0.00,0.50,0.00}{##1}}}
\@namedef{PY@tok@kr}{\let\PY@bf=\textbf\def\PY@tc##1{\textcolor[rgb]{0.00,0.50,0.00}{##1}}}
\@namedef{PY@tok@bp}{\def\PY@tc##1{\textcolor[rgb]{0.00,0.50,0.00}{##1}}}
\@namedef{PY@tok@fm}{\def\PY@tc##1{\textcolor[rgb]{0.00,0.00,1.00}{##1}}}
\@namedef{PY@tok@vc}{\def\PY@tc##1{\textcolor[rgb]{0.10,0.09,0.49}{##1}}}
\@namedef{PY@tok@vg}{\def\PY@tc##1{\textcolor[rgb]{0.10,0.09,0.49}{##1}}}
\@namedef{PY@tok@vi}{\def\PY@tc##1{\textcolor[rgb]{0.10,0.09,0.49}{##1}}}
\@namedef{PY@tok@vm}{\def\PY@tc##1{\textcolor[rgb]{0.10,0.09,0.49}{##1}}}
\@namedef{PY@tok@sa}{\def\PY@tc##1{\textcolor[rgb]{0.73,0.13,0.13}{##1}}}
\@namedef{PY@tok@sb}{\def\PY@tc##1{\textcolor[rgb]{0.73,0.13,0.13}{##1}}}
\@namedef{PY@tok@sc}{\def\PY@tc##1{\textcolor[rgb]{0.73,0.13,0.13}{##1}}}
\@namedef{PY@tok@dl}{\def\PY@tc##1{\textcolor[rgb]{0.73,0.13,0.13}{##1}}}
\@namedef{PY@tok@s2}{\def\PY@tc##1{\textcolor[rgb]{0.73,0.13,0.13}{##1}}}
\@namedef{PY@tok@sh}{\def\PY@tc##1{\textcolor[rgb]{0.73,0.13,0.13}{##1}}}
\@namedef{PY@tok@s1}{\def\PY@tc##1{\textcolor[rgb]{0.73,0.13,0.13}{##1}}}
\@namedef{PY@tok@mb}{\def\PY@tc##1{\textcolor[rgb]{0.40,0.40,0.40}{##1}}}
\@namedef{PY@tok@mf}{\def\PY@tc##1{\textcolor[rgb]{0.40,0.40,0.40}{##1}}}
\@namedef{PY@tok@mh}{\def\PY@tc##1{\textcolor[rgb]{0.40,0.40,0.40}{##1}}}
\@namedef{PY@tok@mi}{\def\PY@tc##1{\textcolor[rgb]{0.40,0.40,0.40}{##1}}}
\@namedef{PY@tok@il}{\def\PY@tc##1{\textcolor[rgb]{0.40,0.40,0.40}{##1}}}
\@namedef{PY@tok@mo}{\def\PY@tc##1{\textcolor[rgb]{0.40,0.40,0.40}{##1}}}
\@namedef{PY@tok@ch}{\let\PY@it=\textit\def\PY@tc##1{\textcolor[rgb]{0.25,0.50,0.50}{##1}}}
\@namedef{PY@tok@cm}{\let\PY@it=\textit\def\PY@tc##1{\textcolor[rgb]{0.25,0.50,0.50}{##1}}}
\@namedef{PY@tok@cpf}{\let\PY@it=\textit\def\PY@tc##1{\textcolor[rgb]{0.25,0.50,0.50}{##1}}}
\@namedef{PY@tok@c1}{\let\PY@it=\textit\def\PY@tc##1{\textcolor[rgb]{0.25,0.50,0.50}{##1}}}
\@namedef{PY@tok@cs}{\let\PY@it=\textit\def\PY@tc##1{\textcolor[rgb]{0.25,0.50,0.50}{##1}}}

\def\PYZbs{\char`\\}
\def\PYZus{\char`\_}

\def\PYZsh{\char`\#}

\def\PYZsq{\char`\'}
\def\PYZdq{\char`\"}

\makeatother

    \makeatletter
        \newbox\Wrappedcontinuationbox 
        \newbox\Wrappedvisiblespacebox 
        \newcommand*\Wrappedvisiblespace {\textcolor{red}{\textvisiblespace}} 
        \newcommand*\Wrappedcontinuationsymbol {\textcolor{red}{\llap{\tiny$\m@th\hookrightarrow$}}} 
        \newcommand*\Wrappedcontinuationindent {3ex } 
        \newcommand*\Wrappedafterbreak {\kern\Wrappedcontinuationindent\copy\Wrappedcontinuationbox} 
        \newcommand*\Wrappedbreaksatspecials {%
            \def\PYGZus{\discretionary{\char`\_}{\Wrappedafterbreak}{\char`\_}}%
            \def\PYGZob{\discretionary{}{\Wrappedafterbreak\char`\{}{\char`\{}}%
            \def\PYGZcb{\discretionary{\char`\}}{\Wrappedafterbreak}{\char`\}}}%
            \def\PYGZca{\discretionary{\char`\^}{\Wrappedafterbreak}{\char`\^}}%
            \def\PYGZam{\discretionary{\char`\&}{\Wrappedafterbreak}{\char`\&}}%
            \def\PYGZlt{\discretionary{}{\Wrappedafterbreak\char`\<}{\char`\<}}%
            \def\PYGZgt{\discretionary{\char`\>}{\Wrappedafterbreak}{\char`\>}}%
            \def\PYGZsh{\discretionary{}{\Wrappedafterbreak\char`\#}{\char`\#}}%
            \def\PYGZpc{\discretionary{}{\Wrappedafterbreak\char`\%}{\char`\%}}%
            \def\PYGZdl{\discretionary{}{\Wrappedafterbreak\char`\$}{\char`\$}}%
            \def\PYGZhy{\discretionary{\char`\-}{\Wrappedafterbreak}{\char`\-}}%
            \def\PYGZsq{\discretionary{}{\Wrappedafterbreak\textquotesingle}{\textquotesingle}}%
            \def\PYGZdq{\discretionary{}{\Wrappedafterbreak\char`\"}{\char`\"}}%
            \def\PYGZti{\discretionary{\char`\~}{\Wrappedafterbreak}{\char`\~}}%
        } 
        \newcommand*\Wrappedbreaksatpunct {%
            \lccode`\~`\.\lowercase{\def~}{\discretionary{\hbox{\char`\.}}{\Wrappedafterbreak}{\hbox{\char`\.}}}%
            \lccode`\~`\,\lowercase{\def~}{\discretionary{\hbox{\char`\,}}{\Wrappedafterbreak}{\hbox{\char`\,}}}%
            \lccode`\~`\;\lowercase{\def~}{\discretionary{\hbox{\char`\;}}{\Wrappedafterbreak}{\hbox{\char`\;}}}%
            \lccode`\~`\:\lowercase{\def~}{\discretionary{\hbox{\char`\:}}{\Wrappedafterbreak}{\hbox{\char`\:}}}%
            \lccode`\~`\?\lowercase{\def~}{\discretionary{\hbox{\char`\?}}{\Wrappedafterbreak}{\hbox{\char`\?}}}%
            \lccode`\~`\!\lowercase{\def~}{\discretionary{\hbox{\char`\!}}{\Wrappedafterbreak}{\hbox{\char`\!}}}%
            \lccode`\~`\/\lowercase{\def~}{\discretionary{\hbox{\char`\/}}{\Wrappedafterbreak}{\hbox{\char`\/}}}%
            \catcode`\.\active
            \catcode`\,\active 
            \catcode`\;\active
            \catcode`\:\active
            \catcode`\?\active
            \catcode`\!\active
            \catcode`\/\active 
            \lccode`\~`\~ 	
        }
    \makeatother

    \let\OriginalVerbatim=\Verbatim
    \makeatletter
    \renewcommand{\Verbatim}[1][1]{%
        \sbox\Wrappedcontinuationbox {\Wrappedcontinuationsymbol}%
        \sbox\Wrappedvisiblespacebox {\FV@SetupFont\Wrappedvisiblespace}%
        \def\FancyVerbFormatLine ##1{\hsize\linewidth
            \vtop{\raggedright\hyphenpenalty\z@\exhyphenpenalty\z@
                \doublehyphendemerits\z@\finalhyphendemerits\z@
                \strut ##1\strut}%
        }%
        \def\FV@Space {%
            \nobreak\hskip\z@ plus\fontdimen3\font minus\fontdimen4\font
            \discretionary{\copy\Wrappedvisiblespacebox}{\Wrappedafterbreak}
            {\kern\fontdimen2\font}%
        }%
        
        \Wrappedbreaksatspecials
        \OriginalVerbatim[#1,codes*=\Wrappedbreaksatpunct]%
    }
    \makeatother

    \definecolor{incolor}{HTML}{303F9F}
    \definecolor{outcolor}{HTML}{D84315}
    \definecolor{cellborder}{HTML}{CFCFCF}
    \definecolor{cellbackground}{HTML}{F7F7F7}
    
    \makeatletter
    \newcommand{\boxspacing}{\kern\kvtcb@left@rule\kern\kvtcb@boxsep}
    \makeatother

    \hypersetup{
      breaklinks=true,  
      colorlinks=true,
      urlcolor=urlcolor,
      linkcolor=linkcolor,
      citecolor=citecolor,
      }
    
\title{\textit{LegalNLP} - Natural Language Processing methods for the Brazilian Legal Language}

\author{Felipe Maia Polo\inst{1}, Gabriel Caiaffa Floriano Mendonça\inst{2}, \\ Kauê Capellato J. Parreira\inst{2}, Lucka Gianvechio\inst{2}, Peterson Cordeiro\inst{2}, \\ Jonathan Batista Ferreira\inst{3}, Leticia Maria Paz de Lima\inst{3}, \\ Antônio Carlos do Amaral Maia\inst{4}, Renato Vicente\inst{2,5}}

\address{Department of Statistics, University of Michigan\\
Ann Arbor, Michigan, United States of America
\nextinstitute
Institute of Mathematics and Statistics, University of São Paulo\\
São Paulo, São Paulo, Brazil
\nextinstitute
Molecular Sciences, University of São Paulo\\
São Paulo, São Paulo, Brazil
\nextinstitute
Tikal Tech\\
São Paulo, São Paulo, Brazil
\nextinstitute
Latam Datalab Serasa Experian\\
São Paulo, São Paulo, Brazil
    \email{maiapolo@umich.edu}
    \email{\{gabrielcaiaffa, kauecapellato, luckagg, peterson.cordeiro, } 
    \email{jonathanbf, leticiamaria, rvicente\}@usp.br}
    \email{antonio.maia@tikal.tech}
}

\begin{document} 

\maketitle

\begin{abstract}
  We present and make available pre-trained language models (Phraser, Word2Vec, Doc2Vec, FastText, and BERT) for the Brazilian legal language, a Python package with functions to facilitate their use, and a set of demonstrations/tutorials containing some applications involving them. Given that our material is built upon legal texts coming from several Brazilian courts, this initiative is extremely helpful for the Brazilian legal field, which lacks other open and specific tools and language models. Our main objective is to catalyze the use of natural language processing tools for legal texts analysis by the Brazilian industry, government, and academia, providing the necessary tools and accessible material.
\end{abstract}

\section{Introduction}

The term Natural Language Processing (NLP) defines the area of research and applications represented by statistical models and algorithms responsible for the analysis and representation of natural language, both phonetic and written. In several countries, the applications of NLP methods in Law are becoming more and more present and represents an increasingly promising future. Classification of legal documents, named entity recognition in legal texts or predictions of legal decisions are just some examples of possible applications.


The first objective of this work is to present and make available pre-trained language models for the Brazilian legal language in addition to a Python package with functions to facilitate the use of these models. The second objective of this work is to present and make available demonstrations\footnote{Full demonstrations are in Google Colab notebooks and are available in our GitHub repository \url{https://github.com/felipemaiapolo/legalnlp}.}, which are accessible to the general public, on how to use language models to analyze Brazilian legal texts in real situations using our models. 


The organization of this work is as follows: in Sections \ref{sec:related} and \ref{sec:context}, we give an overview on NLP methods and mention some related works, besides discussing the importance of this work for the current Brazilian legal context; in Section \ref{sec:tools}, we briefly present the functions that are available in the first version of our package; in Section, \ref{sec:data} we detail the datasets used for training the language models; in Section \ref{sec:models} we present each of the pre-trained language models provided by us and, finally, in Section \ref{sec:demo}, we present two demonstrations involving our models.

The GitHub repository containing our library (models + package) can accessed through \url{https://github.com/felipemaiapolo/legalnlp}.

\section{Related Work}\label{sec:related}

\subsection{NLP background}
Natural Language Processing (NLP) went through a revolution in the last ten years with the advances of Deep Learning methods. Perhaps the first big breakthrough of NLP in the last ten years was the popularization of dense word embeddings with the Word2Vec model \cite{mikolov2013efficient, mikolov2013distributed}. In the way Word2Vec was formulated, the vectors representing words carry the words' meanings derived from the context in which each word can be found. A year later, Doc2Vec was introduced \cite{le2014distributed} as a generalization of Word2Vec to whole texts and, in 2017, FastText \cite{bojanowski2017enriching} was introduced as a way to obtain more robust word vectors, taking into account also word morphology. Also in 2017, the use of self-attention mecanisms and the Transformer architecture \cite{vaswani2017attention} in text processing shifted to whole field of NLP. In more recent years, BERT \cite{devlin2018bert} and derived models used the Transformer architecture to elevate the state of the art in a variety of NLP tasks such as text classification, question answering, and named entity recognition. With the development of new research and methods in NLP, the availability of models for the general public becomes necessary, and then efforts begin to emerge in different countries and fields to develop their own tools.

\subsection{NLP for Brazilian Portuguese and Legal applications}
In the last years there were some efforts to pre-train language models for the Brazilian Portuguese \cite{hartmann2017portuguese,souza2020bertimbau} regarding the more traditional word embeddings (Word2Vec, FastText, etc) as well as BERT models. In spite of that, those models were trained with general documents and were not designed to represent Brazilian legal language. The legal language is unique and demands its own specific language models in order to solve NLP problems that arise everyday in the legal field. In fact, there is a rapid increase in the use of NLP tools to solve real problems in Law, such as identifying the parties in legal proceedings \cite{nguyen2018recurrent}, classification of legal documents according to their administrative labels \cite{dadocument,braz2018document,polo2020predicting} or predicting the area that a lawsuit belongs to \cite{sulea2017exploring}. These works used both classical machine learning paradigm (e.g., TFIDF features + classifier) \cite{polo2020predicting, sulea2017exploring} or deep learning methods \cite{nguyen2018recurrent,dadocument,braz2018document,polo2020predicting} in order to accomplish their goals. Therefore, making pre-trained language models for the Brazilian legal language available can be a turning point for the field in Brazil during the next decades.

\section{Brazilian legal context and the relevance of this work}\label{sec:context}

Law is a challenging field for NLP applications, especially in Brazil. The difficulties are located in the complexity of communication, because legal discipline is fully based on a proper and singular language, spoken for centuries by a minority, and that resembles but does not coincide with the way ordinary people usually speak. Also, Brazilian judicial landscape adds complexity to this communication system, because Brazilian jurisdiction is bigger than Canada's, Australia's, England's, Wales', Scotland's, Ireland's, Belgium's, Switzerland's and France's, combined. The country covers more than 8 millions square kilometers and there are more than 210 million people living in 26 states, each one with their own regional and federal courts, in addition to the national ones. Regionalism can be an obstacle for NLP solutions as there are different ways to speak Portuguese in different regions of the country even within the legal system. For example, judges for different parts of the country can express themselves using different words to designate the same legal reality.

Given the context of the Brazilian legal language, it is necessary (i) for the legal field to have its own pre-trained language models, in order to properly capture singularities of that reality, and (ii) that those models should be trained using texts from several sources, including different courts and Brazilian regions. Given that our library is built upon and for legal texts, from a diversity of sources, this initiative of making functions and models available is a real help in the Brazilian NLP law landscape. It provides researchers in academia, industry or government the tools with necessary context embarked, eliminating or at least mitigating the burden and costs for their activities, permitting them to approach their issues promptly.

\section{Tools for text manipulation}\label{sec:tools}

\subsection{Text cleaning functions}\label{sec:clean}

Our package offers two text cleaning functions, the uses of which are optional. The first one (\textit{clean}) is designed for general use and is compatible with the Phraser, Word2Vec, FastText, and Doc2Vec models presented in Section \ref{sec:models}. In addition, it uses regular expression methods (RegEx) that can extract from texts certain entities such as numbers, values, dates, lawyers IDs\footnote{\textit{Known in Brazil as "código OAB"}.} etc. On the other hand, the second function (\textit{clean\_bert}) is much simpler and makes fewer changes to the texts. We recommend using this function, or some adaptation of it, when using BERTikal, which is our BERT-Base (cased) model pre-trained for the Brazilian legal language.

For more information on cleaning functions, please check our library documentation at \url{https://github.com/felipemaiapolo/legalnlp}.



\subsection{Text tokenization}

For tokenization purposes, we offer two pre-trained Phraser models which can be used in conjunction with the \textit{split} method in Python. Our Phraser models helps us to identify which tokens should be merged (into bigrams, trigrams, and quadrigrams) during tokenization step and are compatible with our Word2Vec, FastText, and Doc2Vec models. More details on the Phraser models is presented in Section \ref{sec:phraser}. Furthermore, BERTikal has its own vocabulary which must be used for tokenization.

\section{Data used to pre-train language models}\label{sec:data}
In total, we used four text datasets. In the process of training Phraser, Word2Vec, Doc2Vec, and FasText we used only the first dataset (Data 1) and for training BERTikal\footnote{Our BERT-Base (cased) model pre-trained for the Brazilian legal language.} we used the other three datasets (Data 2, 3, 4) and started training from the checkpoint provided by a recent work \cite{souza2020bertimbau}. This difference is arbitrary and depended on the context in which the models were trained.

The first and last datasets (Data 1 and 4) are composed exclusively of publications (\textit{publicações}), also known as clippings (\textit{recortes}), from several Brazilian courts for the years 2019 and 2020. On the other hand, the second dataset (Data 2) is composed of longer legal documents mainly from the Court of Justice of São Paulo (TJSP). This is a sample of the dataset used in a master thesis work \cite{massoni2021analise} and was kindly provided to us by its author. The third dataset (Data 3) is composed exclusively of motions (\textit{movimentações}) from several Brazilian courts for the years 2019 and 2020.

In Table \ref{tab:data}, we have information about the number of texts and the approximate size, in gigabytes, of each of the text datasets:




\begin{table}[H]
  \centering
  \caption{Number of texts/documents in each dataset and their approximate size} 
    \begin{tabular}{c|c|c}
    \hline
     Data sources & Number of texts & Approximated size (GB) \\
    \hline
    Data 1 & 1772351 & 3,14 \\
    Data 2 & 34369  & 0,74 \\
    Data 3 & 5246350 & 0,77 \\
    Data 4 & 705521 & 1,12 \\
    \hline
    \end{tabular}%
  \label{tab:data}%
\end{table}%

In  order  to  bring  more  information  about  each  of  the  datasets,  we  detail  their sources. Unfortunately, given the way in which we obtained the data, it was not possible to directly know which courts each of the texts came from.  Therefore, for each of the datasets, we separated a random sample of a maximum of 50k texts and searched for the court of origin for their IDs (Número Único de Processo - NUP)  that  appeared  in  the  body  of  the  text. In  cases  where  more  than  one NUP appeared in the texts, we assumed that the first one was the correct one. Unfortunately, not all texts consulted had a NUP present: for Data 1, 49,006 texts had a NUP; for Data 2, 34,175 texts had a NUP; for Data 3, 1,034 texts had a NUP; finally, for Data 4, 48,802 texts had a NUP. In Table 2, we can see the courts of origin of the texts with their respective relative frequencies, calculated from the random sample already described. We show the fifteen courts with the highest number of texts:


\begin{table}[H]
  \centering
  \caption{Datasets' sources. We show the fifteen courts with the highest number of texts present in each dataset}
    \begin{tabular}{cc|cc|cc|cc}
    \hline 
     \multicolumn{2}{c}{Data 1} & \multicolumn{2}{c}{Data 2} & \multicolumn{2}{c}{Data 3} & \multicolumn{2}{c}{Data 4} \\ 
    \hline
    Tribunal   & \% & Tribunal   & \% & Tribunal   & \% & Tribunal   & \% \\
    \hline
     TJSP   & 36,05 & TJSP   & 99,67 & TJSP   & 86,80 & TJSP   & 43,20 \\
    
    TRT15  & 10,83 & TRT82  & 0,05 & TJAL   & 5,87 & TRT02  & 6,36 \\
    
    TRT02  & 8,13 & TJRJ   & 0,04 & TJRN   & 1,66 & TJBA   & 5,19 \\
    
    TRT03  & 4,00 & TRE82  & 0,03 & TJSC   & 1,08 & TRT15  & 4,47 \\
    
    TJBA   & 3,86 & TJMG   & 0,03 & TJRJ   & 0,78 & TRT01  & 3,72 \\
    
    TRT01  & 3,58 & TRF82  & 0,03 & TJMG   & 0,59 & TRT09  & 2,78 \\
    
    TJSC   & 3,03 & TJCE   & 0,02 & TRT15  & 0,49 & TJSC   & 2,45 \\
    
    TRT04  & 2,94 & TJRS   & 0,02 & TRF03  & 0,49 & TJMS   & 2,39 \\
    
    TRT09  & 2,50 & TJPR   & 0,02 & TJBA   & 0,49 & TRT03  & 2,36 \\
    
    TRT05  & 1,99 & TJBA   & 0,01 & TJMS   & 0,39 & TRF03  & 2,35 \\
    
    TRF03  & 1,88 & TJGO   & 0,01 & TRT02  & 0,20 & TRT05  & 2,24 \\
    
    TJMS   & 1,83 & TJMS   & 0,01 & TRT03  & 0,20 & TJDF   & 2,13 \\
    
    TJDF   & 1,81 & TRE55  & 0,01 & TJAC   & 0,20 & TJSE   & 1,97 \\
    
    TRT12  & 1,47 & TJSC   & 0,01 & TRT01  & 0,20 & TRT04  & 1,90 \\
    
    TJRJ   & 1,41 & TRE17  & 0,01 & TRF02  & 0,10 & TJRJ   & 1,83 \\
    
    
    
    
    
    
    \hline
    \end{tabular}%
  \label{tab:data2}%
\end{table}%
\section{Pre-trained language models}\label{sec:models}

In this section, we go deeper into our pre-trained models. Our focus is to give more details about the models we are making available, also going through the parameters used for the training phase.


\subsection{Phraser}\label{sec:phraser}

Phraser is a statistical method proposed in the natural language processing literature \cite{mikolov2013distributed} for identifying which words, when they appear together, can be considered as unique tokens. This method application is able to identify the relevance of the occurrence of a bigram against the occurrence of the words that make it up separately. Thus, we can identify that a bigram like "São Paulo" should be treated as a single token, for example. If the method is applied a second time in sequence, we can check which are the relevant trigrams and quadrigrams. Since the two applications should be done with different Phraser models, it can be the case that the second application identifies bigrams that were not identified by the first model.

As based on Gensim's package\footnote{See \url{https://radimrehurek.com/gensim_3.8.3/models/phrases.html} - Accessed on 07/30/2021. \label{note1}} for the Python language, we used two Phraser models, which should be applied sequentially. The first model trains for bigrams and the second trains for trigrams, quadrigrams and also finds bigrams that were not found by the first model - an example is given in Section \ref{6 usage example}. The version of the package that was used to train the models is the 3.8.3, though our models can be used with other versions. In case there is any problem resulting from a conflict of versions, we suggest that the user uses the same version as us.



\subsubsection{Data, text preprocessing and models' parameters}

The data used for training the two models are the Data 1 presented in Section \ref{sec:data}. The text preprocessing phase, performed before training the models, was composed by text cleaning and lower casing steps, performed using the general cleaning function presented in Section \ref{sec:clean}. Finally, as model training parameters, we kept the default configuration proposed by the Gensim package, as can be seen in its webpage\footref{note1}.  It is worthy noting that we did not remove stopwords.


\subsubsection{Usage example} \label{6 usage example}

In order to illustrate how Phraser works in practice, we extracted a piece of text from our data. First, we pass the snippet through our general cleaning function, then we tokenize the text, and apply the Phraser models. Then we put the tokens together to reconstruct the texts. Phraser models combine different tokens into unique tokens using the "\_" character:


\begin{itemize}
    \item \textbf{Clean text}: “(...) direito do consumidor origem : bangu regional xxix juizado especial civel ação : [processo] - - recte : fundo de investimento em direitos creditórios (...)"
    \item \textbf{Applying Phraser 1x}: “(...) direito do consumidor origem : bangu regional xxix juizado\_especial civel\_ação : [processo] - - recte : fundo de investimento em direitos\_creditórios (...)"
    \item \textbf{Applying Phraser 2x}: “(...) direito do consumidor origem : bangu\_regional xxix juizado\_especial\_civel\_ação : [processo] - - recte : fundo de investimento em direitos\_creditórios (...)"
\end{itemize}

It is possible to see that the first Phraser model merged the tokens “juizado"~“especial"~resulting in “juizado\_especial"~and “civel" ~“ação" resulting in “civel\_ação", for example. The second Phraser model merged “juizado\_especial"~“civel\_ação" into “juizado\_especial\_civel\_ação" and “bangu"~“regional"~into “bangu\_regional".



\subsection{Word2Vec/Doc2Vec}

Our first models for generating vector representation for tokens and texts (embeddings) are variations of the Word2Vec \cite{mikolov2013distributed,mikolov2013efficient} and Doc2Vec \cite{le2014distributed} methods. In short, the Word2Vec methods generate embeddings for tokens\footnote{Most of the time they are words, n-grams or punctuation.} and that somehow capture the meaning of the various textual elements, based on the contexts in which these elements appear. Doc2Vec methods are extensions/modifications of Word2Vec for generating whole text representations.

The Word2Vec and Doc2Vec methods are presented together in this section as they were trained together using the Gensim\footnote{See \url{https://radimrehurek.com/gensim_3.8.3/models/doc2vec.html} - Accessed on 07/30/2021.}  Python package. The Gensim version used to train the models is 3.8.3, though our models can be used with other versions. In case there is any problem resulting from a conflict of versions, we suggest that the user uses the same version as us.



\subsubsection{Data, text preprocessing and models' parameters}

The dataset used for training the two set of models are the "Data 1"~presented in Section \ref{sec:data}. The text preprocessing phase, performed before training the models, was composed by a text cleaning and lower casing, performed using the general cleaning function presented in Section \ref{sec:clean}, and by tokenization alongside with the application of the Phraser models, described in Section \ref{sec:phraser}, twice in sequence. In that way, we train representations for unigrams, bigrams, trigrams, and quadrigrams.

Both Word2Vec and Doc2Vec were trained with three different sizes (100, 200, 300), with window equal to 15, number of epochs equal to 20 and using its two most known versions: Skip-Gram (SG) and Continuous Bag of Words (CBOW) for Word2Vec \cite{mikolov2013distributed,mikolov2013efficient} and Distributed Memory (DM) and Distributed Bag of Words (DBOW) for Doc2Vec \cite{le2014distributed}. Furthermore, we used the option \textit{dm\_mean}=1 when we trained Doc2Vec DM and Word2Vec CBOW, ignored tokens that appear less than 50 times in our corpus and set all the other parameters to Gensim's default values.

\subsection{FastText}

The FastText \cite{bojanowski2017enriching} methods, like Word2Vec, form a class of models for creating vector representations (embeddings) for tokens. Unlike Word2Vec, which disregards the morphology of the tokens and allocates a different vector for each one of them, the FastText methods consider that each one of the tokens is formed by n-grams of characters or substrings. In this way, the representation of tokens which do not appear in the training set can be inferred from the representation of substrings. Also, rare tokens can have more robust representations than those returned by the Word2Vec methods.

The implementation used to train our FastText models is from the Gensim\footnote{\url{https://radimrehurek.com/gensim/models/fasttext.html} - Accessed on 07/30/2021.} Python package. The Gensim version used to train the models is 4.0.1, though our models can be used with other versions. In case there is any problem resulting from a conflict of versions, we suggest that the user uses the same version as us.



\subsubsection{Data, text preprocessing and models' parameters}

The dataset used for training the two set of models are the "Data 1" presented in Section \ref{sec:data}. The text preprocessing phase, performed before training the models, was composed by a text cleaning and lower casing, performed using the general cleaning function presented in Section \ref{sec:clean}, and by tokenization alongside with the application of the Phraser models, described in Section \ref{sec:phraser}, twice in sequence. In that way, we train representations for unigrams, bigrams, trigrams and quadrigrams.

The FastText methods were trained in the same way as Word2Vec and Doc2Vec.




\subsection{BERTikal}

We call BERTikal our BERT-Base model \cite{devlin2018bert} (cased) for Brazilian legal language. BERT models are models based on neural network architectures called Transformers, which in turn are based on the concept of self-attention \cite{vaswani2017attention}. BERT models are trained with large sets of texts using the self-supervised paradigm, which is basically solving unsupervised problems using supervised techniques. A pre-trained BERT model is capable of generating representations for entire texts and can be adapted for a supervised task, e.g., text classification or question answering, using the fine-tuning mechanism. Fine-tuning consists of training a model that solves a supervised learning task using labeled data and the pre-trained BERT model as its starting point \cite{devlin2018bert}.

BERTikal was trained using the Python package Transformers\footnote{See \url{https://huggingface.co/transformers/} - Accessed on 07/30/2021.} from the company Hugging Face in its 4.2.2 version and its checkpoint made available by us is compatible with PyTorch\footnote{See \url{https://pytorch.org/} - Accessed on 07/30/2021.} 1.9.0. Although we expose the versions of both packages, more current versions can be used in applications of the model, as long as there are no relevant version conflicts.



\subsubsection{Data, text preprocessing and model parameters}

The datasets used for training the model are Data 2, 3, 4 presented in Section \ref{sec:data}. Moreover, the text preprocessing phase, carried out before the training of the models, was composed of a specific text cleaning for BERtikal, performed using the \textit{clean\_bert} function presented in Section \ref{sec:clean}. That function makes few changes to the texts and keeps the uppercased letters.

Our model was trained from the checkpoint made available in Neuralmind's Github repository\footnote{\url{https://github.com/neuralmind-ai/portuguese-bert} - Accessed on 07/30/2021.} by the authors of a recent research \cite{souza2020bertimbau}. In the training phase, we (i) kept the configuration of the model and vocabulary used by the authors, (ii) used the Masked Language Model (MLM) objective with masking probability 0.15, (iii) used one epoch, (iv) batch size equals to 4 texts, and (v) made use of a Tesla T4 GPU. The optimizer settings have been set as the default for the Transformers package from the company Hugging Face\footnote{See \url{https://github.com/huggingface/transformers/blob/master/src/transformers/training_args.py} - Accessed on 10/22/2020.} and the full training took approximately one week to be completed.

\section{Demonstrations}\label{sec:demo}

Now we present some demonstrations of uses of the pre-trained models. The full demos are available on GitHub\footnote{See \url{https://github.com/felipemaiapolo/legalnlp}} and are intended to help the practitioner to make real applications of the models and functions that we make available. The first demonstration presented below consists of an illustration of the vector representations for tokens returned by our 100-sized Word2Vec CBOW model; Secondly, we present solutions to a classification problem, carried out on a legal dataset used in the literature \cite{polo2020predicting} and also made available on Kaggle\footnote{See \url{https://www.kaggle.com/felipepolo/ brazilian-legal-proceedings}.}. We should clarify that the purpose of those classification experiments are not to extensively compare models' performances, but only show an example of real application.




\subsection{Visualizing tokens}

\begin{figure}
    \centering
    \includegraphics[width=1\textwidth]{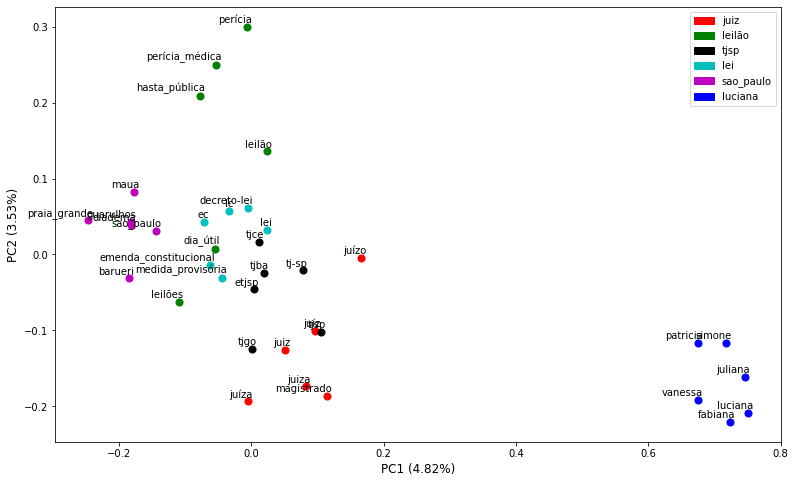}
    \caption{Bidimensional representation of tokens generated by Word2Vec + PCA. The legend shows the central word, the tokens of the same color are the 5 ones with the greatest cosine similarity between their embedding and the central word. You can see how similar words tend to cluster together.}
    \label{fig:EMBEDDINGS_GRAPH}
\end{figure}

In this first example, we perform a dimensionality reduction step on the embeddings returned by our Word2Vec model. To reduce the dimension of the embeddings, we apply principal component analysis (PCA), reducing the dimension from 100 to 2. In this way, it is possible to see the vectors in a bidimensional scatter plot. From there, we draw the graph in Figure \ref{fig:EMBEDDINGS_GRAPH}.


This kind of visualization is useful to exemplify the property of tokens to approximate other similar tokens, that is, those used in similar contexts. In Figure \ref{fig:EMBEDDINGS_GRAPH}, the dots of the same color are the tokens with the greatest cosine similarity between their embedding and the tokens' embeddings in the legend. We can observe that the tokens form clusters. Female names, city names, tokens related to "juiz" (judge, in English) are clustered together, for example.




\subsection{Predicting legal proceedings status}


In this case study, we use the dataset "\textit{Brazilian Legal Proceedings}"\footnote{See \url{https://www.kaggle.com/felipepolo/brazilian-legal-proceedings}.} which has data from 6449 brazillian legal proceedings, each one classified as archived (47.14\%), active (45.23\%) or suspended (7.63\%). Our objective in this study is to, using the most recent text in each proceeding, perform a classification task. This is in contrast with previous work, which the five last texts were used \cite{polo2020predicting}.

To that end, we have combined different language models and classifiers. Firstly, we use a Convolutional Neural Network (CNN) architecture with frozen embbedings (100-sized Word2Vec CBOW models) from our library and from the NILC/USP repository \cite{hartmann2017portuguese}. Secondly, we combined CatBoost \cite{prokhorenkova2017catboost} with our 100-sized Doc2Vec DM. Finally, we combined CatBoost with BERTikal and BERTimbau \cite{souza2020bertimbau} (cased BERT-base) embeddings for texts. We did not fine tune the BERT models.

In order to train and test the different approaches, we randomly split our dataset in training ($70\%$) and test ($30\%$) sets. To train the CNNs, we used batch learning with batch size equals 500, 50 epochs, and an early stop with 15 rounds of patience and a validation split of $10\%$ from training set, we have used 32 filters with kernel size equals 3. For the CatBoost classifier, we used the CatBoost package default hyperparameters' values and an early stop with 100 rounds of patience and a validation split of $10\%$ from training set.

The experiments' results obtained can be observed in Table \ref{tab:metricas_testes}. 

\begin{table}[htbp]
  \centering
  \caption{Results obtained for each model in the test set (score $\pm$ bootstrap std. error)}
    \begin{tabular}{c|c|c|c|c}
    \hline
    Model & Classifier & Accuracy & F1 (Macro Avg.) & F1 (Weighted Avg.)  \\
    \hline
    \hline
    Ours & --- & --- & --- & --- \\
    \hline
    Word2Vec & CNN & $0.84 \pm 0.01$ & $0.80 \pm 0.01$ & $0.84 \pm 0.01$\\
    Doc2Vec & CatBoost & $0.86 \pm 0.01$ & $0.82\pm 0.01$ &$0.85\pm 0.01$  \\
     BERTikal & CatBoost & $0.86 \pm 0.01$ & $0.82 \pm 0.01$ & $0.86 \pm 0.01$\\
    \hline
    \hline
     NILC/USP & --- & --- & --- & --- \\
    \hline
    Word2Vec & CNN & $0.85 \pm 0.01$ & $0.82 \pm 0.01$ & $0.85 \pm 0.01$\\
    \hline
    \hline
    BERTimbau & --- & --- & --- & --- \\
    \hline
    BERT-Base & CatBoost & $0.84 \pm 0.01$ & $0.79 \pm 0.01$ & $0.84 \pm 0.01$  \\
    \hline
    \end{tabular}%
  \label{tab:metricas_testes}%
\end{table}%

From the results presented in Table \ref{tab:metricas_testes} we can see that our models got similar results\footnote{The error bars give us an idea on how results can fluctuate.} at accuracy and F1 scores with the two benchmark models. This is somewhat expected given the dataset we used. As seen in previous work, different approaches have similar results in classification tasks using this dataset \cite{polo2020predicting}. That can happen because classification is not a complex task in this dataset and the presence of some specific words say a lot about the actual class of a document \cite{polo2020predicting}. As this experiment was not intended to show which model performs best in different scenarios, but only give an example of application, we point that direction as a possible future path to go. Future work can be done in order to better understand how our models compare to alternatives in different tasks and datasets.

\section{Conclusion}

In this work, we presented and made available pre-trained language models, functions and demonstrations specific for the Brazilian legal language and field. Our main objective is to catalyze the use of natural language processing tools for legal texts analysis by the Brazilian industry, government and academia, providing the needed tools and accessible material.

\bibliographystyle{sbc}
\bibliography{sbc-template}

\end{document}